\pdfoutput=1

\documentclass[11pt]{article}

\usepackage[preprint]{acl}
\usepackage{hyperref}
\usepackage{url}
\usepackage{graphicx}
\usepackage{booktabs}
\usepackage{multirow}
\usepackage{times}
\usepackage{svg}
\usepackage{listings}
\usepackage{booktabs}
\usepackage{makecell}
\usepackage{latexsym}

\usepackage[T1]{fontenc}

\usepackage[utf8]{inputenc}

\usepackage{microtype}

\usepackage{inconsolata}

\usepackage{graphicx}
\title{Robust Planning with Compound LLM Architectures: \\An LLM-Modulo Approach}

\author{
   Atharva Gundawar \,\,
   Karthik Valmeekam \,\,
   Mudit Verma\thanks{Work done while a PhD student at ASU, now at Google DeepMind}   \,\,
   Subbarao Kambhampati \\
   School of Computing and AI, Arizona State University\,\,
    \\\texttt{\{agundawa, kvalmeek, muditverma, rao\}@asu.edu}
}

\begin{document}
\maketitle
\begin{abstract}
Previous work has attempted to boost Large Language Model (LLM) performance on planning and scheduling tasks through a variety of prompt engineering techniques. While these methods can work within the distributions tested, they are neither robust nor predictable. This limitation can be addressed through compound LLM architectures where LLMs work in conjunction with other components to ensure reliability. In this paper, we present a technical evaluation of a compound LLM architecture--the LLM-Modulo framework.
In this framework, an LLM is paired with a complete set of sound verifiers that validate its output, re-prompting it if it fails. This approach ensures that the system can never output \textit{any} fallacious output, and therefore that every output generated is guaranteed correct--something previous techniques have not been able to claim. Our results, evaluated across four scheduling domains, demonstrate significant performance gains with the LLM-Modulo framework using various models. Additionally, we explore modifications to the base configuration of the framework and assess their impact on overall system performance.

\end{abstract}

\section{Introduction}
As Large Language Models (LLMs) are deployed in contexts far beyond the traditional text completion tasks they were initially trained for, it becomes ever more important to guarantee the correctness of their outputs. Since their release, interest has surged in evaluating LLM capabilities on planning and reasoning tasks, realms traditionally reserved for System 2 cognitive competencies \cite{kahneman2011thinking}. Despite their apparent versatility, studies show that LLMs are not capable of robust reasoning or planning \cite{valmeekam2022large, valmeekam2023planning, silver2022pddl} and the research community is still exploring effective strategies to leverage these models for solving complex planning and reasoning tasks. Even recent advances in prompt engineering that claim performance improvements \cite{wei2022chain, yao2022react} have been shown to not robustly generalize \cite{verma2024brittle, stechly2024chain}. 

Although these studies present evidence that LLMs do not possess System 2 capabilities themselves, their uncanny ability to generate ideas across a wide spectrum of domains--albeit guarantees--can be leveraged to solve a wide variety of System 2 tasks. LLMs can be seen as candidate solution generators within a compound architecture \cite{compound-ai-blog} where other components can vet the solutions and thus ensure that the solutions outputted by the overall system have correctness guarantees where possible.

One such compound architecture is the LLM-Modulo Framework \cite{kambhampati2024llms}. It proposes a conceptual framework whereby an LLM is augmented with a suite of external verifiers and other components which evaluate its proposed answers before deciding whether they should be output. Although the original introduction of the framework \cite{kambhampati2024llms} offered a promising conceptual foundation, it was a position paper and thus lacked a compelling empirical evaluation to demonstrate the framework's viability. 

In this paper, we set out to perform a thorough analysis of the effectiveness of the LLM-Modulo framework in solving complex real-world scheduling tasks. We evaluate the framework on four domains drawn from two benchmarks; Travel Planner \cite{xie2024travelplanner} and three domains from Natural Plan \cite{zheng2024natural}--Trip Planning, Meeting Planning and Calendar Scheduling. Although these domains are more accurately described as scheduling problems, they still fall within the broader scope of planning, with scheduling being a specialized subset of planning. Scheduling problems can be effectively modeled as constraint satisfaction problems (CSPs), which are generally NP-complete. While this complexity is not as high as the PSPACE-complete nature of classical planning \cite{bylander1994computational}, CSPs still present significant challenges for language models. For instance, even when prompted with various popular techniques, such as ReAct \cite{yao2022react} and Chain of Thought \cite{wei2022chain}, LLMs score only a few percent (<5\%) in Travel Planner--whereas humans can manage 100\% accuracy \cite{xie2024travelplanner}.

These benchmarks provide a set of user queries in natural language that represent scheduling tasks for the LLM to solve, as well as code for automatically evaluating LLM responses. We analyze and implement the various components defined in the LLM-Modulo framework and apply the resulting system to these domains. In this generate-test scheduling paradigm, the LLM first proposes a solution which is then verified by a panel of critics. If the critics deem the solution correct, the system outputs it; otherwise, the LLM is prompted again (potentially with feedback) to propose another solution. This interaction loop plays out either until all the critics sign off on the proposed solution or if the loop exceeds a specified maximum budget. LLMs, along with being the solution generators, also play the role of a critic extractor to implement the model-based critics that evaluate the LLM's generated plans and provide critique to the LLM on how to resolve errors. 

While state-of-the-art LLMs (like GPT-4o \cite{openai2024hello} and Claude-3.5-Sonnet \cite{anthropic2024claude}) by themselves show poor performance on these domains, our evaluation shows that placing them within the LLM-Modulo Framework leads to notable improvements. In Travel Planner, the accuracy of GPT-4o improves from 8.3\% to 23.89\% and Claude-3.5-Sonnet improves from 4.4\% to 25\%. Performance gains are significantly higher in the Natural Plan domains; GPT-4o improves from 3.43\% to 40\%! Most importantly, these performances come with \textit{guarantees}: every solution that the LLM-Modulo system produces is deemed correct by the critics.

In the rest of this paper, we first review related work, then provide background on the LLM-Modulo Framework and its operationalization within the chosen domains. Finally, we provide the evaluation of the framework on these domains and analyze the effect of various modifications to the base configuration of the LLM-Modulo Framework on the overall performance.
\begin{figure*}[ht]
    \centering
    \includegraphics[width=0.8\linewidth]{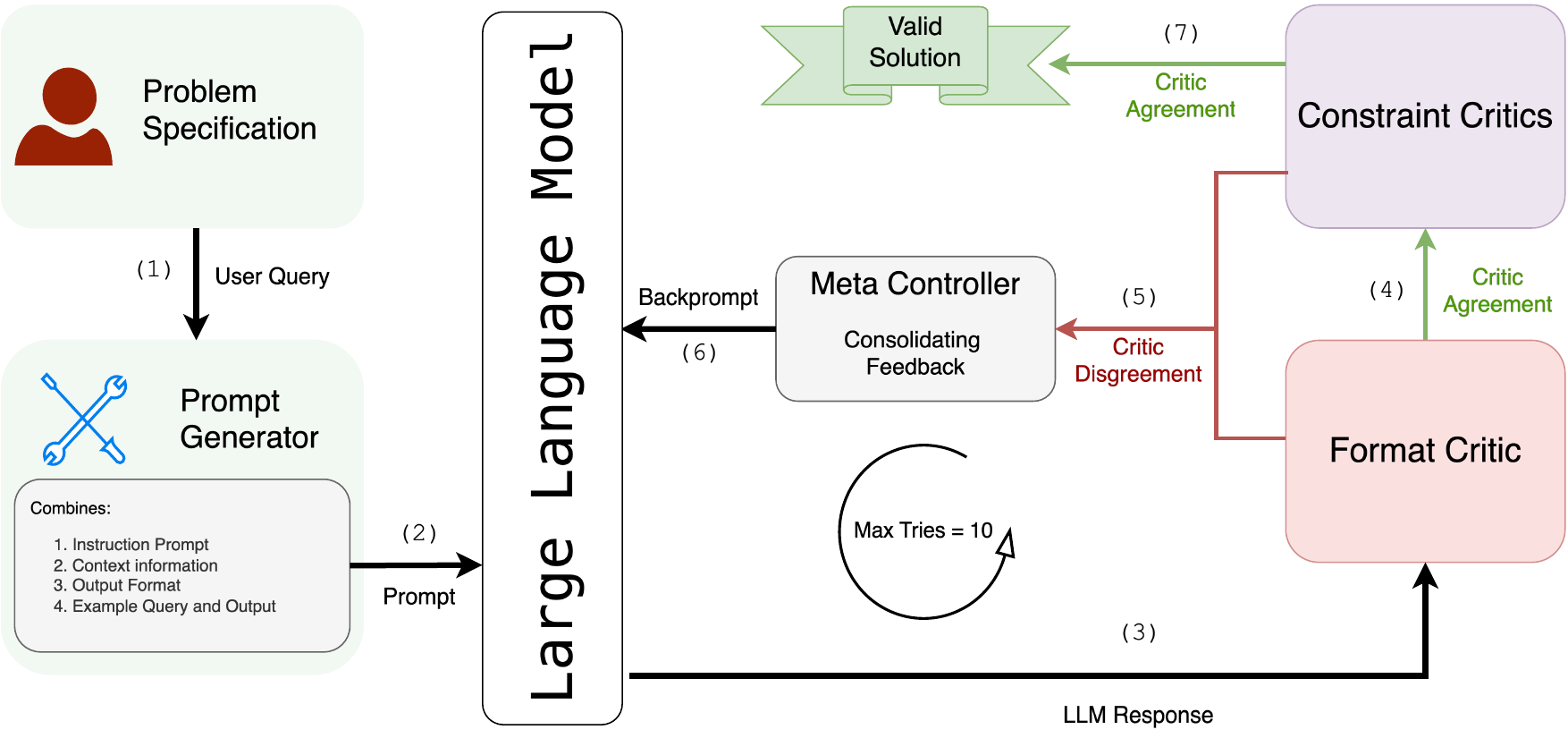}
    \caption{LLM-Modulo Framework for Scheduling domains. The loop begins with (1) generating the prompt using the problem specification by the prompt generator. Then (2) the prompt is passed to the LLM which returns a response. (3) The LLM response is sent to the format critic and if the format critic agrees then (4) it is sent to the constraint critics for checking validity. (5) If any critic disapproves of the response we send the critic feedback to the metacontroller. Then (6) the metacontroller consolidates the critics' evaluations and backprompts the LLM. (7) If all the critics approve of the response the framework returns it as a valid solution}
    \label{fig:llm_modulo_TP}
\end{figure*}
\section{Related Work}
Large Language Models (LLMs) have garnered significant attention due to their versatility in generating seemingly coherent completions across various domains. As a result, researchers have explored using these monolithic models for planning and scheduling tasks, with some even claiming emergent planning capabilities \cite{huang2022inner,bairi2024codeplan,yao2022react,shinn2024reflexion,song2023llm}. However, recent systematic investigations challenged these claims by showing that LLMs are not able to robustly generate executable plans by themselves \cite{valmeekam2024llms, liu2023llm+, silver2022pddl}. Furthermore, recent prompt engineering techniques like Chain of thought \cite{wei2022chain} or ReAct \cite{yao2022react}, that claimed performance gains, have also been shown to lack robust generalization within these tasks \cite{stechly2024chain, dziri2024faith, verma2024brittle}. 

These findings have shifted the focus from relying solely on these monolithic models to compound architectures, where LLMs play key roles alongside other components \cite{compound-ai-blog, valmeekam2023planning, kambhampati2024llms,trinh2024solving,gestrin2024nl2plan, liu2023llm+}. 
The LLM-Modulo framework \cite{kambhampati2024llms} introduces a compound LLM architecture where LLMs serve in various constructive roles, such as generating candidate solutions while other components verify these proposed solutions for correctness.

Other compound LLM architectures, in addition to LLM-Modulo, include solvers that work in conjunction with LLMs rather than verifiers \cite{trinh2024solving,gestrin2024nl2plan, liu2023llm+}. However, these approaches are constrained by the expressiveness and computational limitations of the solvers.
Additional approaches modify LLM prompts by including relevant information or filtering out unsafe or erroneous content, thus aiding the generation of better solutions \cite{lewis2020retrieval,han2024torchopera,xu2024p}. Pipelines that optimize prompts relative to specific target metrics have also been developed to maximize system performance \cite{khattab2023dspy}. However, these systems continue to face robustness challenges as they still rely on the LLM’s generations, which do not come with guarantees. The LLM-Modulo framework addresses this issue by incorporating verifiers that validate LLM-generated responses before final output.

The LLM-Modulo framework, as presented in the position paper \cite{kambhampati2024llms}, has not yet undergone extensive empirical evaluation to assess its effectiveness across various tasks. In our work, we aim to address this gap by evaluating a base configuration of the LLM-Modulo framework across four scheduling domains. We also investigate several modifications to the base configuration and analyze their impact on overall performance.

\section{LLM-Modulo Framework}
The underlying architecture of the LLM-Modulo Framework is that of a generate-test-critique loop, with the LLM proposing solutions and a bank of critics critiquing the proposals. The soundness guarantees of the system stem from having sound critics as part of the loop. The completeness of the system depends on the generator i.e., the LLM's ability to generate all potentially relevant candidates.
\begin{table*}[h]
    \begin{tabular}{@{}lcccccccc@{}}
    \toprule
        \multirow{2}{*}{\textbf{Models}} & \multicolumn{2}{c}{\textbf{OSU}} & \multicolumn{2}{c}{\textbf{TP}} & \multicolumn{2}{c}{\textbf{MP}} & \multicolumn{2}{c}{\textbf{CS}} \\ \cmidrule(l){2-9} 
                                         & \textbf{Direct}   & \textbf{LM}  & \textbf{Direct}  & \textbf{LM}  & \textbf{Direct}  & \textbf{LM}  & \textbf{Direct}  & \textbf{LM}  \\ \midrule
        \textbf{Gpt 4o-mini}             & 2.78\%            & 15.00\%         & 6.18\%           & 12.06\%      & 32.80\%           & 51.90\%      & 36.90\%          & 61.60\%      \\
        \textbf{Gpt 4o}                  & 8.33\%            & 23.89\%      & 3.43\%           & 40.00\%         & 49.10\%           & 59.80\%            & 56.10\%          & 83.30\%      \\

        \textbf{Claude 3.5 Sonnet}       & 4.44\%            & 25.00\%         & 39.43\%          & 47.00\%            & 57.10\%                & 69.5\%            & 72.90\%          & 88.80\%      \\ \bottomrule
    \end{tabular}
    \caption{Performance of various LLMs on Travel Planner (OSU), Meeting Planning (MP), Trip Planning (TP), and Calendar Scheduling (CS).  The \textbf{Direct} column represents direct prompting, while the \textbf{LM} column shows the results of 10 iterations of LLM-Modulo.}
    \label{tab:main_results_table}
\end{table*}
\subsection{Problem Specification}
As shown in Figure \ref{fig:llm_modulo_TP}, the interaction loop starts by providing the problem specification to the LLM. A complete problem specification can be provided by the user or LLMs can be used in conjunction to flesh out an incomplete problem specification before including it in the initial input prompt. In our evaluation, the four domains by design contain all information required to generate a feasible solution. 
\paragraph{Prompt Generator:} We include a prompt generator that takes care of providing an instruction prompt along with relevant context information. Along with this, the prompt generator provides instructions on the output format of the generated plan by providing examples as part of the prompt. For Travel Planner, the prompts are one-shot prompts, as in a single example is provided whereas for the domains in Natural Plan the prompts are five-shot.
\subsection{Critics}
Once the LLM provides a candidate solution the critics within the LLM-Modulo framework can evaluate the candidate over both hard and soft constraints. Hard constraints refer to various correctness verification constraints while soft constraints can include more abstract notions like style, preference conformance, explicability etc. The critics can provide varying levels of feedback ranging from binary to constructive feedback wherein alternatives can be offered.  

\paragraph{Extracting Critics:}
For all the domains, we utilize the code present as part of the benchmark as critics to evaluate the proposed solutions. However, these implementations can be extracted from the LLM itself. We tested this by prompting GPT-4-Turbo with the required contextual information such as the objective of the critic, available tools or databases with corresponding function declaration, and the input plan as JSON (along with the JSON schema) as part of the prompt. We observed that with minimal modifications (such as fixing function call signature and syntax which itself can be automated via critics such as compilers and parsers) the generated critics matched the efficacy of the pre-existing ones. We present the prompts used for the critic extraction in \ref{critic_extract}.

In our implementation (Figure \ref{fig:llm_modulo_TP}), all the responses first go through a format critic before going the other critics. If there are issues at either the format level or the constraints level, the critique is sent to the metacontroller to consolidate a single backprompt and query the LLM again.

\subsection{Backprompt (Meta) Controller}
The metacontroller takes on the control flow if the critics find issues with the solution. It contains the decision-making logic to stitch together the critic responses, choose which backprompts to allow for (if a certain pedagogical prompting is in effect), or other consolidation of various backprompts. The metacontroller interfaces with the LLM and includes other information such as instructions, database context, formatting, and few-shot examples along with the compiled backprompt. 

As shown in Figure \ref{fig:llm_modulo_TP}, the meta-controller first consolidates the format backprompts, if any, and queries the LLM to address the format issues. If no formatting issues are detected, the LLM response is sent to the other critics. If those critics find errors, the meta-controller aggregates the response of those critics and generates a single backprompt for the LLM. This backprompt includes the original prompt, the LLM’s response from the previous iteration, and the response provided by the bank of critics. This interaction loop continues until all of the critics agree to the generated solution or until a specified maximum budget (set to 10 iterations) is exceeded.

\section{Evaluation}
\label{eval}
In this section, we present the empirical results of evaluating the LLM Modulo framework on four specific scheduling domains: Travel Planner, Trip Planning, Meeting Planning, and Calendar Scheduling. Our experiments utilize three models: GPT-4o-mini, GPT-4o, and Claude-3.5-Sonnet. We use temperature 0 across all our experiments. 

\subsection{Travel Planner}
Travel Planner provides a sandboxed environment to generate itineraries for travel plan queries simulated using various datasets, such as flights, restaurants, distances between locations, available transport options, and accommodation choices, to name a few. As mentioned previously, Travel Planner evaluates generated solutions based on hard constraints and common-sense constraints. We use the validation dataset (within the sole-planning mode) consisting of 180 queries for our experiments. The prompts within the sole-planning mode provide necessary context to the LLM instead of expecting the LLM to perform tool use \cite{schick2024toolformer,paranjape2023art,hsieh2023tool}. The query is combined with a system prompt that specifies relevant information, including details about available accommodations, restaurants, attractions, flights, and road transportation options. We prompt the LLM to generate a travel plan in JSON format, which includes, for each day of the trip, the day number, number of travelers, current city, transportation method, and recommendations for breakfast, lunch, dinner, and attractions.

Initially, every LLM-generated response passes through a format critic that validates the JSON structure, ensuring it is parsable and contains all required key-value pairs. Only if the response meets these criteria is it forwarded to the other critics.

As detailed in Table \ref{tab:main_results_table}, the performance of all models improves significantly when integrated within the LLM-Modulo framework. Specifically with 10 iterations of LLM-Modulo, state-of-the-art models like GPT-4o and Claude-3.5-Sonnet see notable increase in performance with 15.56\% and 20.56\% increase respectively. Even smaller models like GPT-4o-mini show accuracy gains, from 2.78\% to 15\%. 

\begin{figure*}[ht]
    \centering
    \includegraphics[width=1\linewidth]{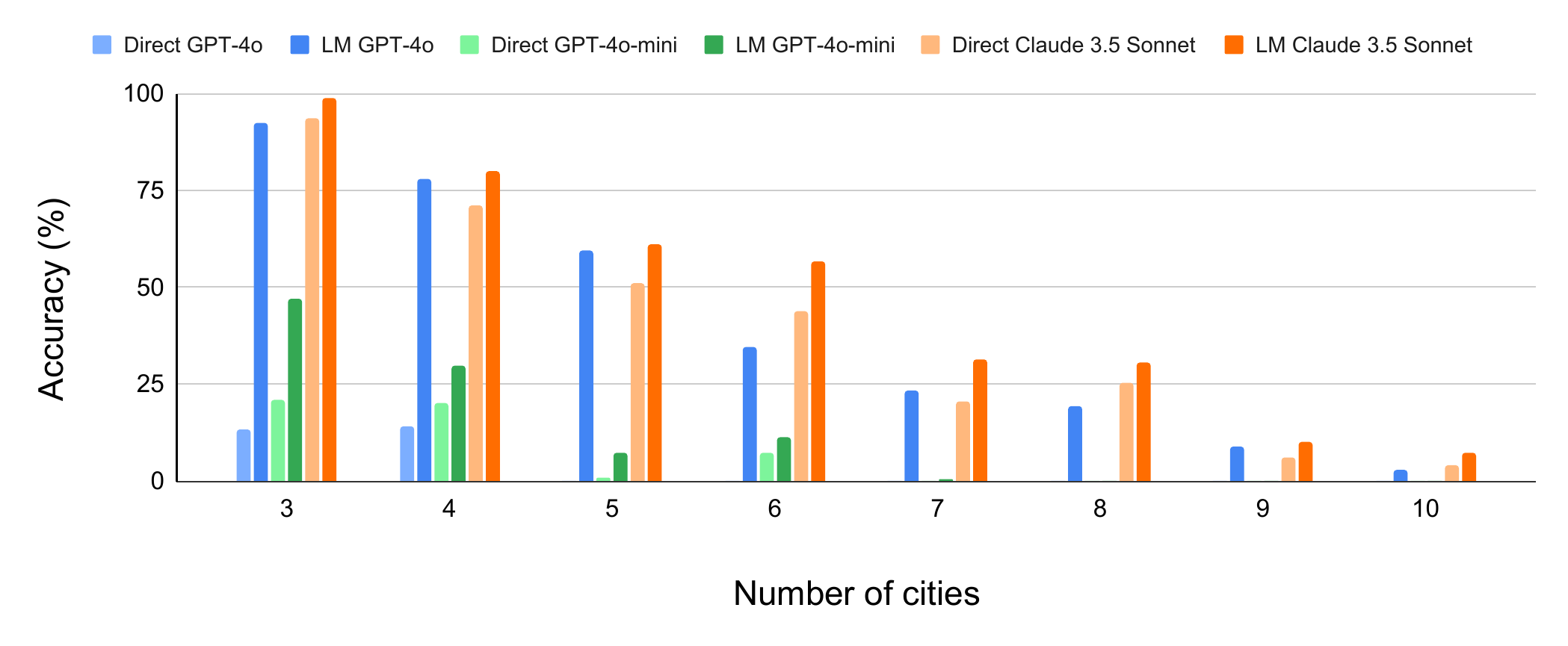}
    \caption{Performance of models with direct prompting (lighter colors) and with LLM Modulo (darker colors) on Trip Planning across subsets with varying complexity. LLM Modulo is indicated by LM.}
    \label{fig:tp_lm}
\end{figure*}

We have also tested popular prompting methods such as CoT and ReAct with GPT-4o-mini. While CoT and React methods achieved 2.2\% and 0\% respectively, LLM-Modulo achieves 15\%.
\subsection{Trip Planning}
The Trip Planning domain comprises 1,600 scenarios, each requiring the formulation of a travel itinerary that adheres to specified constraints. The complexity of the scenarios in this dataset is determined by the number of cities mentioned in the query, which can vary from three to ten. This categorization divides the dataset into eight subsets, each containing 200 instances.

For the first iteration of LLM Modulo, we use the same prompt provided by the authors of Natural Plan, as mentioned earlier. The LLM is instructed to return a regex-parsable response, and we provide 5 examples to form a 5-shot prompt. Similar to the Travel Planner domain, the LLM’s response is first sent to a format critic, which checks whether the response is correctly parsed. If the format is valid, the response is then passed to three additional critics. The first critic verifies that the duration of the stay in each city matches the specified constraints. The second critic checks whether the required flights are available based on the planned itinerary. The third critic ensures that for each event, the traveler is in the correct city during the event’s duration. 

The results, as illustrated in Table \ref{tab:main_results_table}, show significant improvements with LLM-Modulo across models. While GPT-4o-mini improves from 6.18\% to 12.06\% and Claude-3.5-Sonnet improves from 39.43\% to 47\%, with GPT-4o we see a significant jump from 3.43\% to 40\%. 

\subsection{Meeting Planning}
Meeting planning involves 1,000 scenarios, where the objective is to maximize the number of friends met within given time and location constraints, using a distance matrix. 
The dataset is divided into ten subsets, each consisting of 100 instances, categorized by the number of individuals involved, ranging from one to ten participants.

For the first iteration of LLM Modulo, we use the same prompt provided by the authors of Natural Plan, as previously mentioned. The LLM is instructed to generate a regex-parsable response, and we provide five examples, creating a 5-shot prompt. After passing through the format critic, the sole critic checks if any part of the plan conflicts with the meeting schedule or location. This critic verifies whether the meeting is feasible at each step, ensuring that the person the agent is planning to meet is available at the specified time and place, and that the agent is in the correct location according to the plan.

The three models, GPT-4o-mini, GPT-4o, and Claude-3.5-Sonnet show significant performance gains when employing the LLM-Modulo approach compared to direct prompting. Claude-3.5-Sonnet’s accuracy jumps from 57.1\% to 69.5\%, while GPT-4o improves from 49.1\% to 59.8\%. GPT-4o-mini demonstrates the highest improvement, with its accuracy rising from 32.8\% to 51.9\%. 

\subsection{Calendar Scheduling}
The Calendar Scheduling domain encompasses 1,000 scenarios that challenge planners to identify suitable meeting times ranging from 30 minutes to one hour, considering the schedules of \textit{n} individuals over \textit{m} days. The complexity of scenarios is derived from the number of participants (\textit{n}) and the range of days (\textit{m}). The dataset is divided into two subsets: one varying the number of participants from three to seven across a single day, and the other varying the days from one to five with two participants.
For the first iteration of LLM Modulo, we use the same prompt provided by the authors of Natural Plan, as previously mentioned. The LLM is instructed to generate a regex-parsable response, and we provide five examples, creating a 5-shot prompt. Similar to the previous domains, after ensuring that the format is valid, the response is then passed to the only critic which looks if the generated response meeting time clashes with any attendee's schedule.

Similar to the other domains, LLM Modulo framework has enhanced performance within the Calendar Scheduling dataset. Each model tested in this domain achieved a significant boost in performance; smaller models like GPT-4o-mini increased from the original 36.9\% to 61.6\%! GPT-4o and Claude-3.5-Sonnet both experience performance increases, with GPT-4o reaching an accuracy of 83.3\% after 10 iterations and Claude-3.5-Sonnet reaching 88.8\%.
\begin{table}[t]
\centering
\begin{tabular}{llcc}
\toprule
\textbf{Domain} & \textbf{Model} & \textbf{Direct} & \makecell{\textbf{LM}} \\
\midrule
\multirow{3}{*}{\makecell[l]{Travel\\Planner}} & 3.5 Sonnet & 0 & 12 \\
& 4o & 1 & 8 \\
& 4o-mini & 2 & 4 \\
\midrule
\multirow{3}{*}{\makecell[l]{Trip\\Planning}} & 3.5 Sonnet & 8 & 15 \\
& 4o & 0 & 6 \\
& 4o-mini & 0 & 0 \\
\midrule
\multirow{3}{*}{\makecell[l]{Meeting\\Planning}} & 3.5 Sonnet & 11 & 32 \\
& 4o & 8 & 19 \\
& 4o-mini & 0 & 5 \\
\midrule
\multirow{3}{*}{\makecell[l]{Calendar\\Scheduling}} & 3.5 Sonnet & 68 & 87 \\
& 4o & 50 & 75 \\
& 4o-mini & 20 & 36 \\
\bottomrule
\end{tabular}
\caption{Number of correct instances by Direct prompting and 10 iterations of LLM Modulo (LM) on the hardest subsets of the domains. Claude-3.5-Sonnet, GPT-4o, GPT-4o-mini are referred as 3.5-Sonnet, 4o, 4o-mini respectively}
\label{tab:hardest_subsets}
\end{table}
\subsection{Performance on the hardest subsets}
The LLM Modulo approach not only enhances overall performance across all domains but also demonstrates improvements in handling instances of varying complexity within each domain, including the hardest subsets (see Figure \ref{fig:tp_lm}). These hardest subsets are defined by scenarios with the highest complexity in each domain and include a significant number of instances.

The improvements in the number of correct responses for each model on these hardest subsets are shown in Table \ref{tab:hardest_subsets}. 
For Travel Planner, the hardest subset consists of 60 queries that include the maximum number of constraints (three per query). Notably, for this domain, Claude-3.5-Sonnet’s performance increased from 0 to 12 correct responses using LLM Modulo. In Trip Planning, over 200 instances involving the highest number of cities (ten cities), Claude-3.5-Sonnet improved from 8 to 15 correct responses (an increase of 7) while the GPT-4o model went from 0 to 6 correct responses. For Meeting Planning, the hardest subset comprises 100 tasks with the maximum number of people (ten people). Within this subset, Claude-3.5-Sonnet showed the highest improvement--from 11 to 32 correct responses--using LLM Modulo. In Calendar Scheduling, over 100 scenarios that require accounting for the most schedules while planning (two people over five days), GPT-4o’s performance notably increased from 50 to 75 correct responses. 

Overall, these results demonstrate that LLM Modulo substantially boosts performance on the most demanding tasks, even for the largest and most complex inputs.

\section{Exploring Modifications within the LLM-Modulo Framework}
\label{mod}
Previously, in our evaluations, we focused on a basic configuration of the LLM-Modulo framework. The LLM proposes a solution, a panel of critics evaluate it, if the critics find any errors, the metacontroller puts together a backprompt containing the response and the critics' evaluation, and queries the LLM back. This loop runs till the LLM provides the right response or a specified budget is exceeded. Our deliberate focus on this configuration is to establish the viability of the overall system in its most basic form but there could be various other configurations (both domain-dependent and independent) that could have a positive impact on the system's performance. In this section, we look at some of the obvious additions/modifications to this basic configuration and analyze their effects on overall performance.  

We evaluate the following modifications of the LLM-Modulo Framework; more context about previous iterations, filtering unfit values, querying for multiple solutions, types of feedback and prompt engineering. We use GPT-4o-mini for all our experiments as (a) showing improvements in a smaller model can act as a potential lower bound for improvement and (b) the model is cost-effective. 

\subsection{Adding context from previous iterations}

\begin{figure}[ht]
    \centering
    \includegraphics[width=1\linewidth]{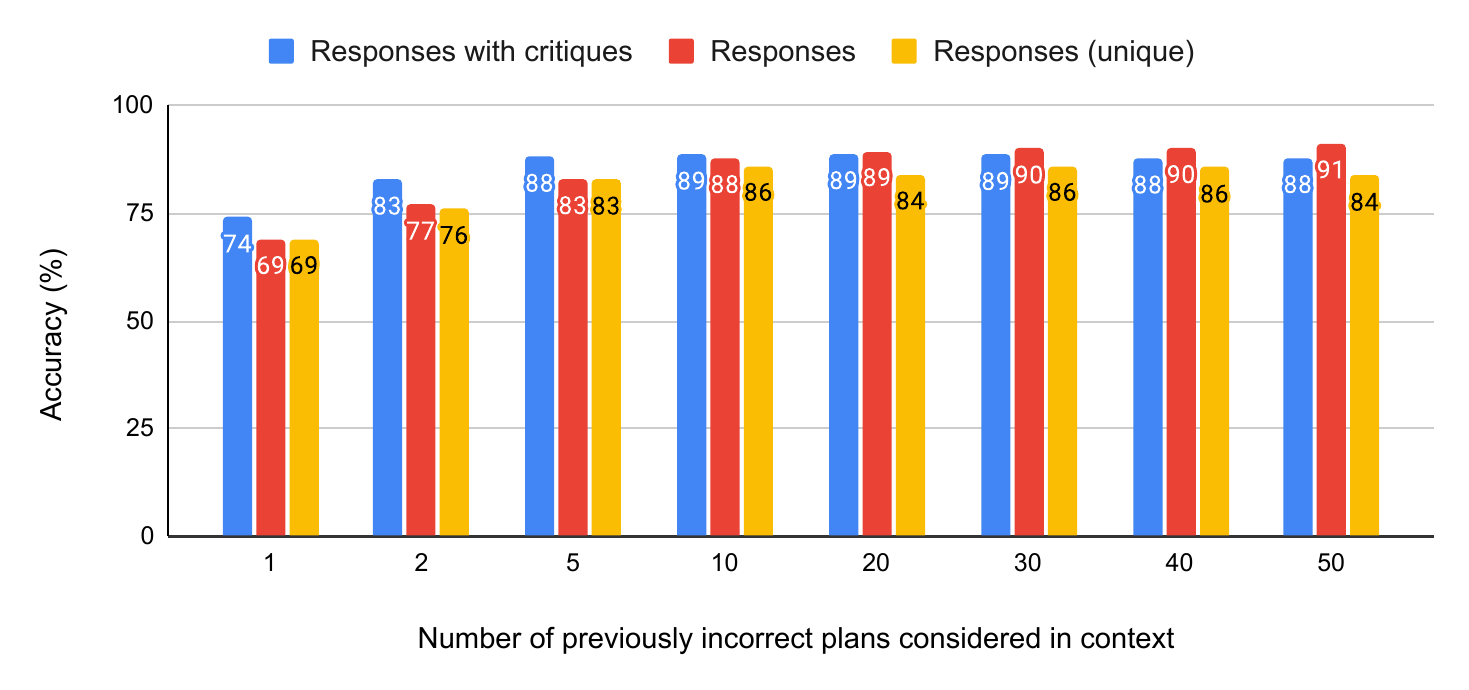}
    \caption{Effect of including context from previous iterations on model performance in the Calendar Scheduling domain using the GPT-4o-mini model. All values are rounded to the nearest integer}
    \label{fig:n_ply_history}
\end{figure}

We investigate the impact of incorporating context from previous generations by including previously generated (incorrect) plans and the feedback on them. To analyze this, we performed two ablations: (1) including only the incorrect plans without their feedback, and (2) including only the \textit{unique} incorrect plans to avoid redundancy. These experiments were conducted in the calendar scheduling domain using the GPT-4o-mini model, with varying values of \textit{n}, representing the number of previously incorrect plans included as context. The context contained the most recent \textit{n} incorrect plans, and each configuration was evaluated over 50 iterations of LLM-Modulo.

Our findings indicate that increasing the number of previously generated incorrect plans (n) as part of the context improves overall accuracy, although gains become marginal beyond a value of 10 (as shown in Figure \ref{fig:n_ply_history}). Within the kind of responses that are part of the backprompt, the configuration that included only unique incorrect plans consistently underperformed compared to the others. Additionally, the inclusion of critiques alongside the incorrect plans did not significantly affect performance.

\subsection{Filtering}
When constraint values are specified in the base prompt, as in the case of Travel Planner, we can iteratively remove values from the backprompts that have been tested in previous LLM Modulo loops and have been deemed unfit by the critics. We employ this filtering strategy by removing values for constraint such as Room Type, Room Rules, and Minimum Nights for accommodations when they are flagged by the accommodation critic.

For example, suppose the LLM selects a "Luxury building studio" in Los Angeles as the accommodation for the first day, which requires a minimum stay of 7 nights. If you plan to stay in Los Angeles for only 5 days, this accommodation will never be valid, and any plan including this accommodation will be invalid. Therefore, in all subsequent iterations of LLM Modulo, we remove this accommodation and its details from the backprompts we send to the LLM.

In terms of performance, applying this filtering strategy slightly increased the success rate for GPT-4o-mini from 12.22\% (LLM Modulo Baseline) to 12.67\% (LLM Modulo with Filtering). For this specific experiment, we also evaluated GPT-4o and we found a significant increase in the performance with this modification; from 21.11\% (LLM Modulo Baseline) to 36.11\% (LLM Modulo with Filtering).

\subsection{Querying for multiple solutions}
We examine diversification at the generation level by prompting the base LLM to generate multiple solutions for each presented query. If the LLM modulo loop were structured as a chain of \( n \) iterations, this strategy would resemble a tree of depth \( n \). Every path from the start node to a leaf node represents a baseline LLM modulo chain. We employ a breadth-first search exploration strategy, with a maximum depth of 10 (or stopping whenever a valid plan is found), for calendar scheduling with GPT-4o-mini, and we report an increase in performance. The accuracy at the first prompt itself is 45.20\%, with an increase of 8.3\%. After 10 iterations, the accuracy of the system reaches 87.40\%, representing an increase of 25.80\%.

\subsection{Types of feedback}
\begin{figure}[ht]
    \centering
    \includegraphics[width=1.0\linewidth]{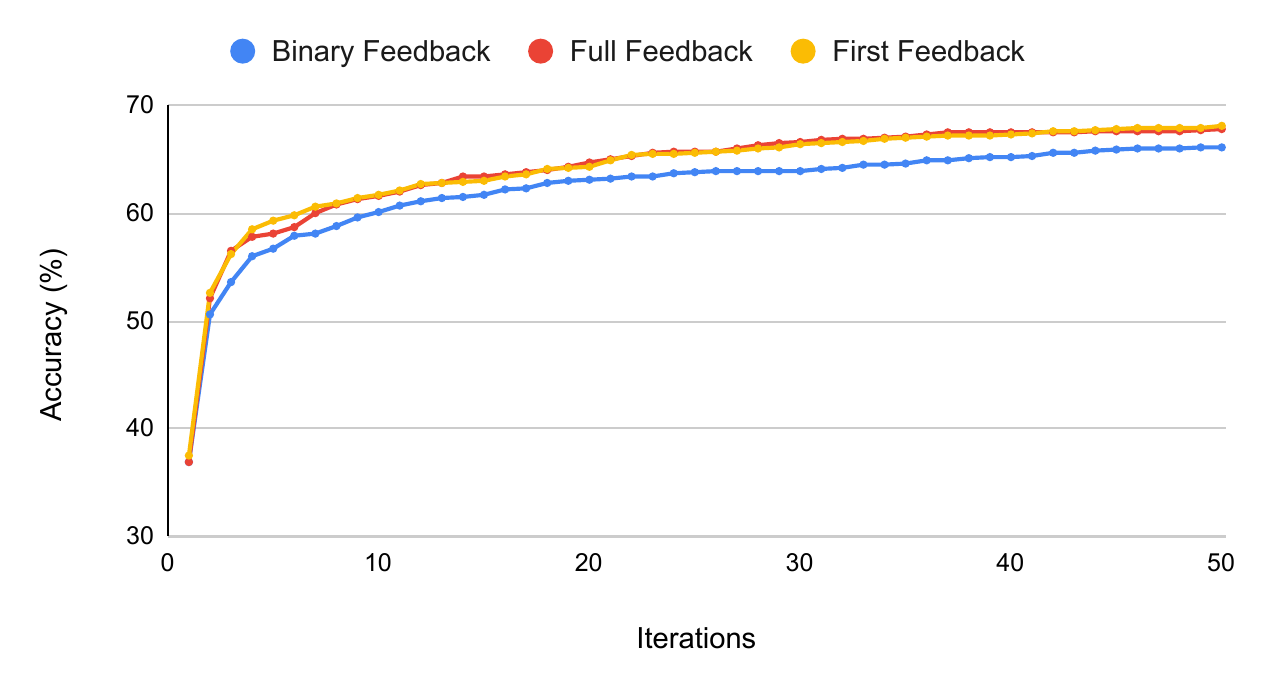}
    \caption{Comparison of Full Feedback, Binary Feedback, and First Feedback across different iterations in the calendar scheduling domain using GPT-4o-mini. Results are reported as performance percentages for each feedback type.}
    \label{fig:feedback_types}
\end{figure}

Thus far, all backprompts to the base model have consisted of a collated list of all critiques or feedback from the generation of the previous iteration. This subsection explores the utility of this feedback method through two experiments. In the first experiment, we utilize binary feedback by prompting the LLM with \textit{"This time doesn't work. Come up with an alternative schedule"} instead of providing the full feedback. In the second experiment, we provide only the first critique to test whether the base LLM becomes overloaded with the volume of feedback it receives.

The results on calendar scheduling with GPT-4o-mini indicate that when backprompted with just binary critiques, the performance of the LLM Modulo is consistently lower compared to when full feedback is provided. However, when backprompted with only the first critique, the performance is comparable to that of presenting all the critiques simultaneously (as shown in Figure \ref{fig:feedback_types}).

\subsection{Prompt engineering}
In this experiment, we examine the impact of a simple prompt engineering technique; zero-shot Chain-of-Thought (COT) \cite{kojima2023largelanguagemodelszeroshot} applied to calendar scheduling. For each prompt, we append "Think step-by-step" at the end and then query the LLM.

\begin{figure}[ht]
    \centering
    \includegraphics[width=1.0\linewidth]{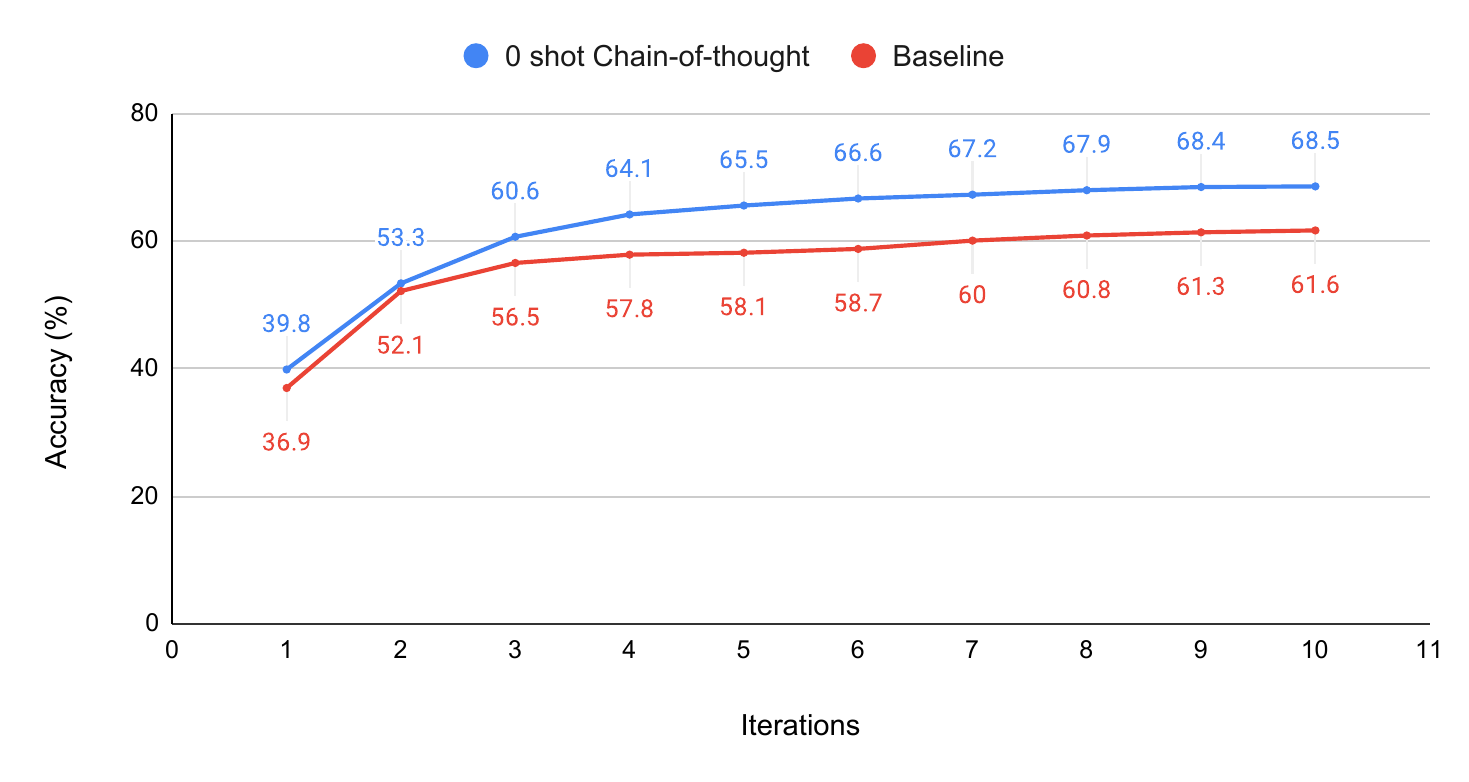}
    \caption{Accuracy improvement of GPT-4o-mini Modulo on calendar scheduling with zero-shot Chain-of-Thought prompting(blue), compared to baseline(red).}
    \label{fig:prompt_perturbation}
\end{figure}

As illustrated in Figure \ref{fig:prompt_perturbation}, the accuracy of GPT-4o-mini Modulo steadily improves with each iteration, reaching 68.5\% at the 10th iteration. This represents a 6.9\% increase over the baseline performance of the LLM-Modulo framework.  

Overall, our preliminary investigations show that while the base configuration of the LLM-Modulo framework already promise significant improvements, these additional modifications can further improve the overall system performance.
\section{Conclusion}

In this paper, we investigated the effectiveness of integrating a large language model (LLM) within a compound architecture, specifically the LLM-Modulo framework, across four complex scheduling tasks. Our findings demonstrate that while state-of-the-art LLMs perform poorly when used in isolation, embedding them within the LLM-Modulo architecture leads to significant performance improvements. We provided a detailed walkthrough of the framework’s instantiation in each scheduling domain and outlined how the various components can be configured effectively including extracting the critics from LLMs itself. Additionally, we explored various modifications to the base LLM-Modulo configuration and analyzed the effect of these modifications on the performance of the overall system. We showed that while some modifications enhanced the performance, some didn't. Overall, we established that LLM-Modulo can provide significant performance gains with guarantees in complex scheduling tasks. 

\section{Limitations}
Our experiments, as described in Section \ref{eval}, were constrained to 10 iterations of the LLM-Modulo framework due to cost limitations. Similarly, in Section \ref{mod}, we restricted our modification experiments to the evaluation of GPT-4o-mini for the same reason. While we assessed the base configuration alongside several modified versions—resulting in significant performance improvements with formal guarantees—our exploration of potential modifications was not exhaustive. Given these constraints, we focused on key adjustments that delivered notable gains, though further exploration could potentially enhance performance. These additional modifications are beyond the scope of the current work and remain an avenue for future investigation.
\bibliography{acl}
\newpage
\onecolumn
\appendix
\section{Appendix}

\subsection{Performance across all subsets}
\begin{figure*}[ht]
    \centering
    \includegraphics[width=1\linewidth]{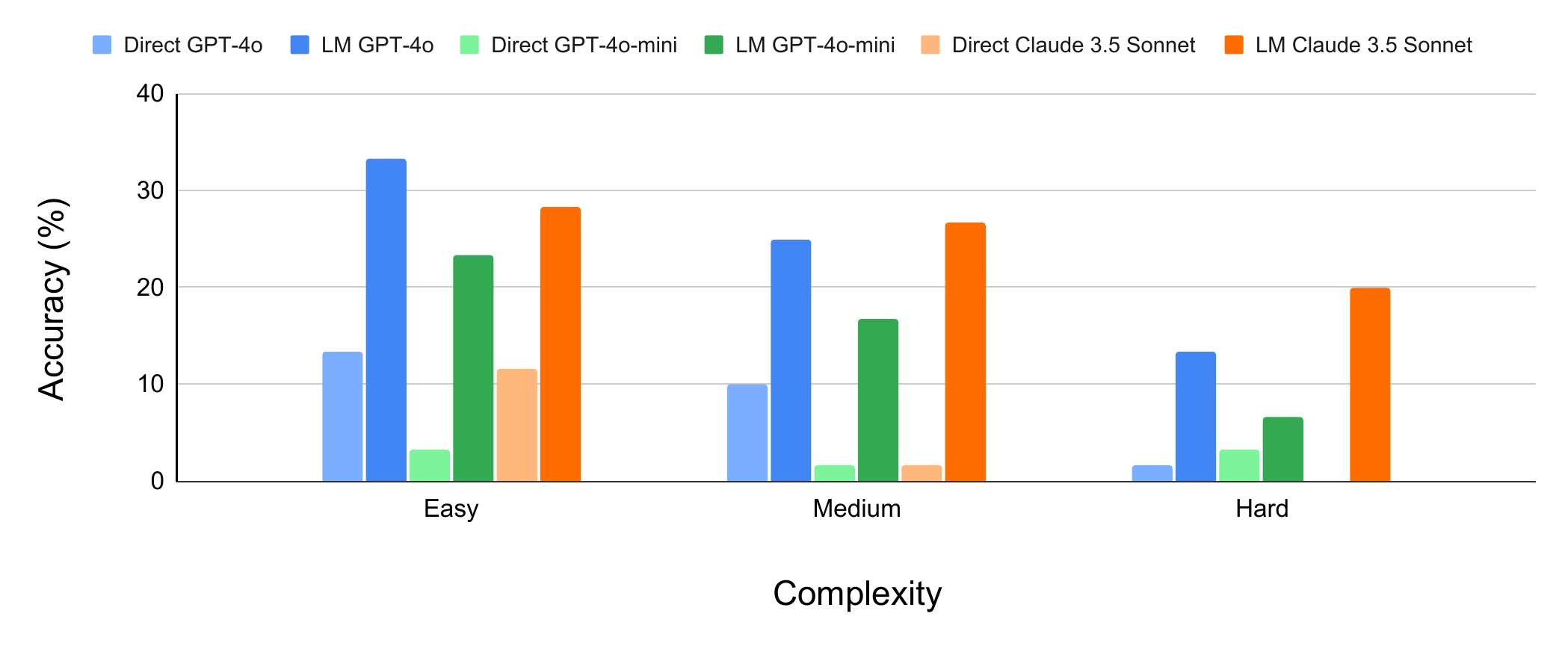}
    \caption{Performance of models with direct prompting (lighter colors) and with LLM Modulo (darker colors) on Travel Planner across all subsets. LLM Modulo is indicated by LM.}
    \label{fig:osu_lm}
\end{figure*}
\begin{figure*}[ht]
    \centering
    \includegraphics[width=1\linewidth]{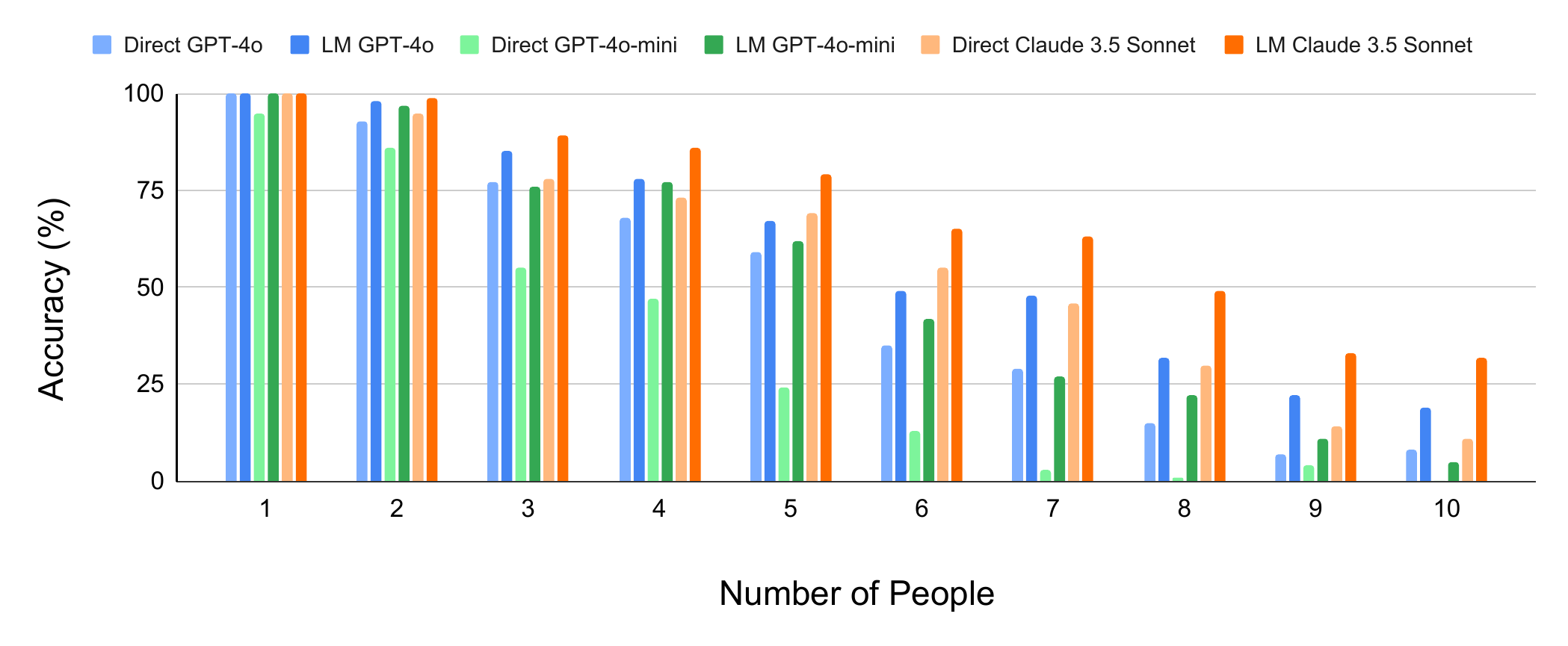}
    \caption{Performance of models with direct prompting (lighter colors) and with LLM Modulo (darker colors) on Meeting Planning across all subsets. LLM Modulo is indicated by LM.}
    \label{fig:mp_lm}
\end{figure*}
\begin{figure*}[ht]
    \centering
    \includegraphics[width=1\linewidth]{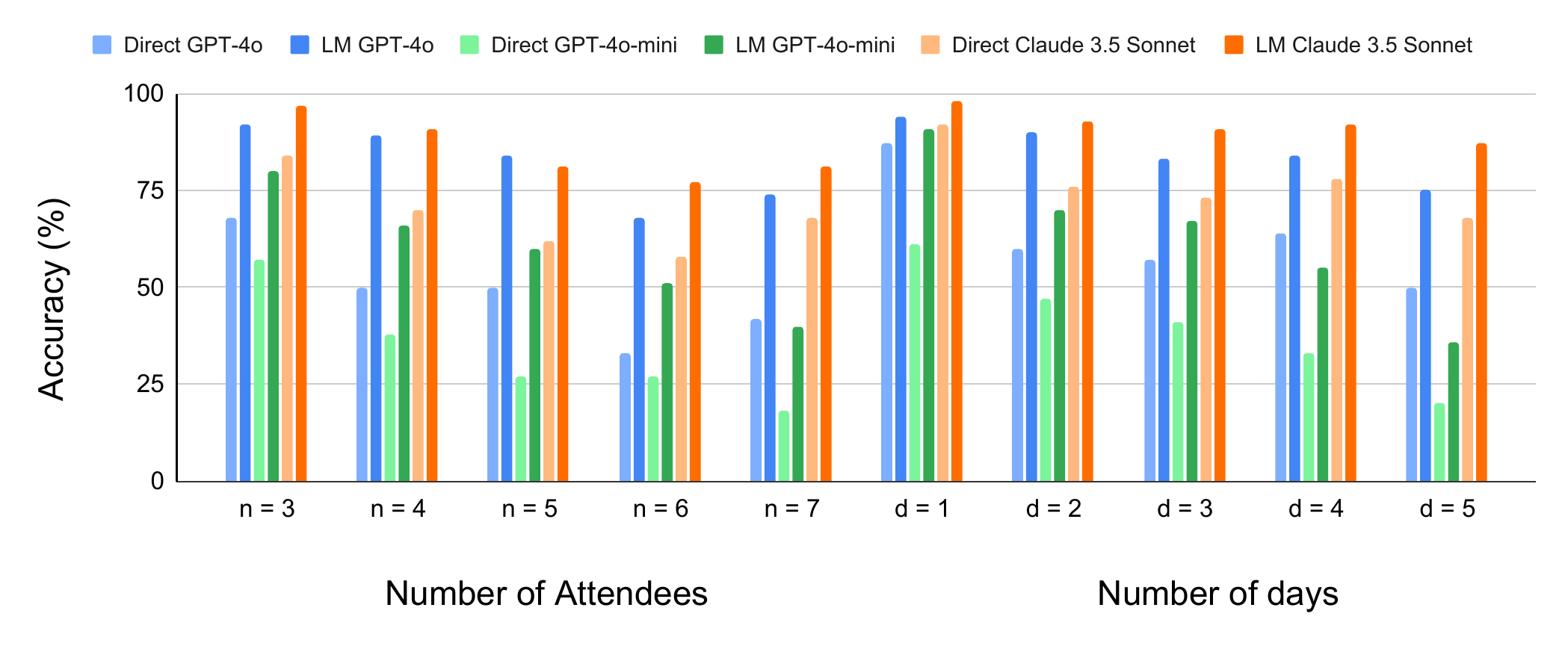}
    \caption{Performance of models with direct prompting (lighter colors) and with LLM Modulo (darker colors) on Calendar Scheduling across all subsets. LLM Modulo is indicated by LM.}
    \label{fig:cs_lm}
\end{figure*}

\newpage

\subsection{Travel Planner}
The complete prompts and templates are available at: \href{https://github.com/Atharva-Gundawar/LLM-Modulo-prompts}{github.com/Atharva-Gundawar/LLM-Modulo-prompts}
\begin{lstlisting}
You are a proficient planner. Based on the provided information and query, please give me a detailed plan, including specifics such as flight numbers (e.g., F0123456), restaurant names, and accommodation names. Note that all the information in your plan should be derived from the provided data. You should give a travel plan in JSON format as shown in the example below. Additionally, all details should align with commonsense. The symbol '-' indicates that information is unnecessary. For example, in the provided sample, you do not need to plan after returning to the departure city. When you travel to two cities in one day, you should note it in the 'current_city' section as in the example (i.e., from A to B).

***** Example *****
Query: Could you create a travel plan for 7 people from Ithaca to Charlotte spanning 3 days, from March 8th to March 14th, 2022, with a budget of $30,200?
Travel Plan:
[
    {
        "day": 1,
        "people_number": 7,
        "current_city": "from Ithaca to Charlotte",
        "transportation": "Flight Number: F3633413, from Ithaca to Charlotte, Departure Time: 05:38, Arrival Time: 07:46",
        "breakfast": "Nagaland's Kitchen, Charlotte",
        "attraction": "The Charlotte Museum of History, Charlotte",
        "lunch": "Cafe Maple Street, Charlotte",
        "dinner": "Bombay Vada Pav, Charlotte",
        "accommodation": "Affordable Spacious Refurbished Room in Bushwick!, Charlotte"
    },
    .
    .
    .
]
***** Example Ends *****

Given information: [{'Description': 'Attractions in Myrtle ...
Query: Please create a travel plan for me where I'll be departing from Washington and heading to Myrtle Beach for a 3-day trip from March 13th to March 15th, 2022. Can you help me keep this journey within a budget of $1,400?
Travel Plan (please only output the JSON string without explanatory information):

==============RESPONSE OF 4o-mini=================

```json
[
    {
        "day": 1,
        "people_number": 1,
        "current_city": "from Washington to Myrtle Beach",
        "transportation": "Flight Number: F3792603, from Washington to Myrtle Beach, Departure Time: 09:19, Arrival Time: 10:59",
        "breakfast": "-",
        "attraction": "SkyWheel Myrtle Beach, Myrtle Beach; WonderWorks Myrtle Beach, Myrtle Beach",
        "lunch": "Exotic India, Myrtle Beach",
        "dinner": "Twigly, Myrtle Beach",
        "accommodation": "Cozy Brooklyn Room - Next to Pratt Institute, Myrtle Beach"
    },
    .
    .
    .
]
```

==============Backprompt=================

You are a proficient planner. Based on the provided information, query, and the backprompt, please fix the given travel plan, including specifics such as flight numbers (e.g., F0123456), restaurant names, and accommodation names. Note that all the information in your plan should be derived from the provided data. You should give a travel plan in JSON format as shown in the example below. Additionally, all details should align with commonsense. The symbol '-' indicates that information is unnecessary. For example, in the provided sample, you do not need to plan after returning to the departure city. When you travel to two cities in one day, you should note it in the 'current_city' section as in the example (i.e., from A to B).

***** Example *****
Query: Could you create a travel plan for 7 people from Ithaca to Charlotte spanning 3 days, from March 8th to March 14th, 2022, with a budget of $30,200?
Travel Plan:
[
    {
        "day": 1,
        "people_number": 7,
        "current_city": "from Ithaca to Charlotte",
        "transportation": "Flight Number: F3633413, from Ithaca to Charlotte, Departure Time: 05:38, Arrival Time: 07:46",
        "breakfast": "Nagaland's Kitchen, Charlotte",
        "attraction": "The Charlotte Museum of History, Charlotte",
        "lunch": "Cafe Maple Street, Charlotte",
        "dinner": "Bombay Vada Pav, Charlotte",
        "accommodation": "Affordable Spacious Refurbished Room in Bushwick!, Charlotte"
    },
    .
    .
    .
]
***** Example Ends *****

Given information: [{'Description': 'Attractions in Myrtle ...
Query: Please create a travel plan for me where I'll be departing from Washington and heading to Myrtle Beach for a 3-day trip from March 13th to March 15th, 2022. Can you help me keep this journey within a budget of $1,400?

Travel Plan:

[
    {
        "day": 1,
        "people_number": 1,
        "current_city": "from Washington to Myrtle Beach",
        "transportation": "Flight Number: F3792603, from Washington to Myrtle Beach, Departure Time: 09:19, Arrival Time: 10:59",
        "breakfast": "-",
        "attraction": "SkyWheel Myrtle Beach, Myrtle Beach; WonderWorks Myrtle Beach, Myrtle Beach",
        "lunch": "Exotic India, Myrtle Beach",
        "dinner": "Twigly, Myrtle Beach",
        "accommodation": "Cozy Brooklyn Room - Next to Pratt Institute, Myrtle Beach"
    },
    .
    .
    .
]

By inspecting your plan we find the following issue. Please refine your plan according to the feedback below:
1. The accommodation Cozy Brooklyn Room - Next to Pratt Institute, Myrtle Beach do not obey the minumum nights rule.
2. The breakfast in day 3 is invalid or not in the data provided.

Fixed Travel Plan (please only output the JSON string without explanatory information):

\end{lstlisting}

\subsection{Natural Plan}
The complete prompts and templates are available at: \href{https://github.com/Atharva-Gundawar/LLM-Modulo-prompts}{github.com/Atharva-Gundawar/LLM-Modulo-prompts}
\subsubsection{Changed Evaluation Function}
Note that for Calendar Scheduling, the domain had assumed that all problems consist of a single solution therefore, the evaluation was a one-to-one mapping between the generated response and a golden solution. We found a discrepancy within the domain as certain problems had multiple valid solutions. Therefore, we modified the evaluation function for this domain to account for \textit{all} the valid solutions for a given instance.

\subsubsection{Trip Planning}
\begin{lstlisting}
You are an expert at planning trips. You are given a few constraints regarding the cities to visit and the durations of staying at each city. You are also given the flight information between the cities('and' suggests flights are possible both ways and 'from - to -' suggests one way flights). You can travel to any city only once.
Produce only one plan, strictly adhering to the format shown in the examples below, without any extra information. Your answer shold start with 'SOLUTION:'.
Here are a few example tasks and solutions:

TASK: You plan to visit 10 European cities for 21 days in total. You only take direct flights to commute between cities. You plan to stay in Vienna for 3 days. You want to spend 5 days in Frankfurt. You want to spend 2 days in Oslo. You are going to attend a wedding in Oslo between day 20 and day 
21. You want to spend 3 days in Prague. You would like to visit Valencia for 2 days. You want to meet a friend in Valencia between day 17 and day 18. You plan to stay in Dubrovnik for 2 days. You would like to visit Edinburgh for 5 days. From day 1 to day 5, there is a annual show you want to attend
in Edinburgh. You plan to stay in London for 2 days. You plan to visit relatives in London between day 12 and day 13. You would like to visit Munich for 3 days. You would like to meet your friends at Munich between day 18 and day 20 to tour together. You would like to visit Budapest for 3 days.

Here are the cities that have direct flights:
Valencia and Munich, Vienna and Munich, Vienna and Valencia, London and Budapest, London and Oslo, Edinburgh and Budapest, Frankfurt and Budapest, Frankfurt and London, Prague and Oslo, Edinburgh and Oslo, Edinburgh and Munich, Prague and Munich, London and Prague, Edinburgh and London, Edinburgh 
and Frankfurt, Dubrovnik and Munich, Dubrovnik and Vienna, Munich and Oslo, Dubrovnik and Oslo, Budapest and Munich, Frankfurt and Prague, Vienna and London, Frankfurt and Vienna, Frankfurt and Oslo, Frankfurt and Munich, Vienna and Oslo, Vienna and Prague, Budapest and Oslo, Budapest and Prague, 
London and Valencia, London and Munich, Frankfurt and Dubrovnik, Prague and Valencia, Frankfurt and Valencia, Edinburgh and Prague, Vienna and Budapest.

Find a trip plan of visiting the cities for 21 days by taking direct flights to commute between them.
SOLUTION: Here is the trip plan for visiting the 10 European cities for 21 days:

**Day 1-5:** Arriving in Edinburgh and visit Edinburgh for 5 days.
**Day 5:** Fly from Edinburgh to Frankfurt.
.
.
.
**Day 20:** Fly from Munich to Oslo.
**Day 20-21:** Visit Oslo for 2 days.

TASK: You ...
.
.
.


Query:
You plan to visit 10 European cities for 25 days in total. You only take direct flights to commute between cities. You plan to stay in Berlin for 4 days. You have to attend a workshop in Berlin between day 12 and day 15. You would like to visit Prague for 2 days. You plan to stay in Stuttgart for 5 
days. You would like to visit Manchester for 3 days. You want to spend 2 days in Nice. You would like to visit Reykjavik for 2 days. You would like to visit Florence for 3 days. You plan to stay in Vilnius for 5 days. You would like to meet your friends at Vilnius between day 15 and day 19 to tour 
together. You plan to stay in Oslo for 4 days. You would like to visit Dubrovnik for 4 days. You plan to visit relatives in Dubrovnik between day 1 and day 4.

Here are the cities that have direct flights:
from Reykjavik to Stuttgart, Manchester and Stuttgart, Nice and Berlin, Oslo and Prague, Stuttgart and Berlin, Manchester and Nice, Reykjavik and Oslo, Reykjavik and Prague, Manchester and Prague, Reykjavik and Berlin, Dubrovnik and Manchester, Manchester and Oslo, Manchester and Berlin, Prague and 
Florence, Berlin and Vilnius, Dubrovnik and Oslo, Nice and Oslo, Berlin and Oslo, Nice and Reykjavik, Vilnius and Oslo.

Find a trip plan of visiting the cities for 25 days by taking direct flights to commute between them.

==============RESPONSE OF 4o-mini=================

SOLUTION: Here is the trip plan for visiting the 10 European cities for 25 days:

**Day 1-4:** Arriving in Dubrovnik and visit Dubrovnik for 4 days.
**Day 4:** Fly from Dubrovnik to Stuttgart.
**Day 4-8:** Visit Stuttgart for 5 days.
**Day 8:** Fly from Stuttgart to Berlin.
**Day 8-12:** Visit Berlin for 4 days.
**Day 12:** Attend workshop in Berlin for 4 days (Day 12-15).
**Day 15:** Fly from Berlin to Vilnius.
**Day 15-19:** Visit Vilnius for 5 days.
**Day 19:** Fly from Vilnius to Oslo.
**Day 19-22:** Visit Oslo for 4 days.
**Day 22:** Fly from Oslo to Nice.
**Day 22-24:** Visit Nice for 2 days.
**Day 24:** Fly from Nice to Manchester.
**Day 24-27:** Visit Manchester for 3 days.
**Day 27:** Fly from Manchester to Prague.
**Day 27-28:** Visit Prague for 2 days.
**Day 28:** Fly from Prague to Florence.
**Day 28-30:** Visit Florence for 3 days.
**Day 30:** Fly from Florence to Reykjavik.
**Day 30-32:** Visit Reykjavik for 2 days.

==============Backprompt=================

You are an expert at planning trips. Fix the below given trip schedule such that it meets as many specified constraints as possible. You are given a few constraints regarding the cities to visit and the durations of staying at each city. You are also given the flight information between the cities('and' suggests flights are possible both ways and 'from - to -' suggests one way flights). You can travel to any city only once.
Produce only one plan, strictly adhering to the format shown in the examples below, without any extra information. Your answer shold start with 'SOLUTION:'.
Here are a few example tasks and solutions:

TASK: You plan to visit 10 European cities for 21 days in total. You only take direct flights to commute between cities. You plan to stay in Vienna for 3 days. You want to spend 5 days in Frankfurt. You want to spend 2 days in Oslo. You are going to attend a wedding in Oslo between day 20 and day 
21. You want to spend 3 days in Prague. You would like to visit Valencia for 2 days. You want to meet a friend in Valencia between day 17 and day 18. You plan to stay in Dubrovnik for 2 days. You would like to visit Edinburgh for 5 days. From day 1 to day 5, there is a annual show you want to attend
in Edinburgh. You plan to stay in London for 2 days. You plan to visit relatives in London between day 12 and day 13. You would like to visit Munich for 3 days. You would like to meet your friends at Munich between day 18 and day 20 to tour together. You would like to visit Budapest for 3 days.

Here are the cities that have direct flights:
Valencia and Munich, Vienna and Munich, Vienna and Valencia, London and Budapest, London and Oslo, Edinburgh and Budapest, Frankfurt and Budapest, Frankfurt and London, Prague and Oslo, Edinburgh and Oslo, Edinburgh and Munich, Prague and Munich, London and Prague, Edinburgh and London, Edinburgh 
and Frankfurt, Dubrovnik and Munich, Dubrovnik and Vienna, Munich and Oslo, Dubrovnik and Oslo, Budapest and Munich, Frankfurt and Prague, Vienna and London, Frankfurt and Vienna, Frankfurt and Oslo, Frankfurt and Munich, Vienna and Oslo, Vienna and Prague, Budapest and Oslo, Budapest and Prague, 
London and Valencia, London and Munich, Frankfurt and Dubrovnik, Prague and Valencia, Frankfurt and Valencia, Edinburgh and Prague, Vienna and Budapest.

Find a trip plan of visiting the cities for 21 days by taking direct flights to commute between them.
SOLUTION: Here is the trip plan for visiting the 10 European cities for 21 days:

**Day 1-5:** Arriving in Edinburgh and visit Edinburgh for 5 days.
**Day 5:** Fly from Edinburgh to Frankfurt.
.
.
.
**Day 20:** Fly from Munich to Oslo.
**Day 20-21:** Visit Oslo for 2 days.

TASK: You ...
.
.
.

Query:
You plan to visit 10 European cities for 25 days in total. You only take direct flights to commute between cities. You plan to stay in Berlin for 4 days. You have to attend a workshop in Berlin between day 12 and day 15. You would like to visit Prague for 2 days. You plan to stay in Stuttgart for 5 
days. You would like to visit Manchester for 3 days. You want to spend 2 days in Nice. You would like to visit Reykjavik for 2 days. You would like to visit Florence for 3 days. You plan to stay in Vilnius for 5 days. You would like to meet your friends at Vilnius between day 15 and day 19 to tour 
together. You plan to stay in Oslo for 4 days. You would like to visit Dubrovnik for 4 days. You plan to visit relatives in Dubrovnik between day 1 and day 4.

Here are the cities that have direct flights:
from Reykjavik to Stuttgart, Manchester and Stuttgart, Nice and Berlin, Oslo and Prague, Stuttgart and Berlin, Manchester and Nice, Reykjavik and Oslo, Reykjavik and Prague, Manchester and Prague, Reykjavik and Berlin, Dubrovnik and Manchester, Manchester and Oslo, Manchester and Berlin, Prague and 
Florence, Berlin and Vilnius, Dubrovnik and Oslo, Nice and Oslo, Berlin and Oslo, Nice and Reykjavik, Vilnius and Oslo.

Find a trip plan of visiting the cities for 25 days by taking direct flights to commute between them.

Incorrect plan:
SOLUTION: Here is the trip plan for visiting the 10 European cities for 25 days:

**Day 1-4:** Arriving in Dubrovnik and visit Dubrovnik for 4 days.
**Day 4:** Fly from Dubrovnik to Stuttgart.
**Day 4-8:** Visit Stuttgart for 5 days.
**Day 8:** Fly from Stuttgart to Berlin.
**Day 8-12:** Visit Berlin for 4 days.
**Day 12:** Attend workshop in Berlin for 4 days (Day 12-15).
**Day 15:** Fly from Berlin to Vilnius.
**Day 15-19:** Visit Vilnius for 5 days.
**Day 19:** Fly from Vilnius to Oslo.
**Day 19-22:** Visit Oslo for 4 days.
**Day 22:** Fly from Oslo to Nice.
**Day 22-24:** Visit Nice for 2 days.
**Day 24:** Fly from Nice to Manchester.
**Day 24-27:** Visit Manchester for 3 days.
**Day 27:** Fly from Manchester to Prague.
**Day 27-28:** Visit Prague for 2 days.
**Day 28:** Fly from Prague to Florence.
**Day 28-30:** Visit Florence for 3 days.
**Day 30:** Fly from Florence to Reykjavik.
**Day 30-32:** Visit Reykjavik for 2 days.

Errors with the above plan:
Total duration of plan is 41, expected 34

Fixed plan:

\end{lstlisting}

\subsubsection{Meeting Planning}
\begin{lstlisting}
You are a meeting planner agent. Generate a meeting schedule that meets as many specified constraints as possible. 
Produce only one plan, strictly adhering to the format shown in the examples below, without any extra information.

You are visiting San Francisco for the day and want to meet as many friends as possible. Solve the problem by considering various different schedules and picking the best one to optimize your goals.

Travel distances (in minutes):
North Beach to Pacific Heights: 8.
North Beach to Golden Gate Park: 22.
North Beach to Sunset District: 27.
North Beach to Chinatown: 6.
Pacific Heights to North Beach: 9.
.
.
.

Chinatown to Sunset District: 29.

CONSTRAINTS: You arrive at North Beach at 9:00AM. Sarah will be at Pacific Heights from 6:45PM to 9:15PM. You'd ...

SOLUTION:You start at North Beach at 9:00AM. You travel to Golden Gate Park in 22 minutes and arrive at 9:22AM. You ...

You are visiting San Francisco for ...

.
.
.

Query:
You are visiting San Francisco for the day and want to meet as many friends as possible. Solve the problem by considering various different schedules and picking the best one to optimize your goals.

Travel distances (in minutes):
Fisherman's Wharf to Embarcadero: 8.
Fisherman's Wharf to Golden Gate Park: 25.
Fisherman's Wharf to Mission District: 22.
Fisherman's Wharf to Union Square: 13.
Embarcadero to Fisherman's Wharf: 6.
.
.
.

CONSTRAINTS: You arrive at Fisherman's Wharf at 9:00AM. Barbara will be at Embarcadero from 12:15PM to 6:45PM. You'd like to meet Barbara for a minimum of 105 minutes. Ashley will be at Golden Gate Park from 9:30AM to 6:00PM. You'd like to meet Ashley for a minimum of 75 minutes. Ronald will be at 
Mission District from 2:30PM to 8:15PM. You'd like to meet Ronald for a minimum of 45 minutes. Robert will be at Union Square from 11:45AM to 9:00PM. You'd like to meet Robert for a minimum of 60 minutes.

==============RESPONSE OF 4o-mini=================

SOLUTION: You start at Fisherman's Wharf at 9:00AM. You travel to Golden Gate Park in 25 minutes and arrive at 9:25AM. You meet Ashley for 75 minutes from 9:25AM to 10:40AM. You travel to Union Square in 22 minutes and arrive at 11:02AM. You wait until 11:45AM. You meet Robert for 60
minutes from 11:45AM to 12:45PM. You travel to Embarcadero in 10 minutes and arrive at 12:55PM. You meet Barbara for 105 minutes from 12:55PM to 2:40PM. You travel to Mission District in 20 minutes and arrive at 3:00PM. You meet Ronald for 45 minutes from 3:00PM to 3:45PM.

==============Backprompt=================

You are a meeting planner agent. Fix the below given meeting schedule such that it meets as many specified constraints as possible. 
Produce only one plan, strictly adhering to the format shown in the examples below, without any extra information.

You are visiting San Francisco for the day and want to meet as many friends as possible. Solve the problem by considering various different schedules and picking the best one to optimize your goals.

Travel distances (in minutes):
North Beach to Pacific Heights: 8.
North Beach to Golden Gate Park: 22.
North Beach to Sunset District: 27.
North Beach to Chinatown: 6.
Pacific Heights to North Beach: 9.
.
.
.

Chinatown to Sunset District: 29.

CONSTRAINTS: You arrive at North Beach at 9:00AM. Sarah will be at Pacific Heights from 6:45PM to 9:15PM. You'd ...

SOLUTION:You start at North Beach at 9:00AM. You travel to Golden Gate Park in 22 minutes and arrive at 9:22AM. You ...

You are visiting San Francisco for ...

.
.
.

Query:
You are visiting San Francisco for the day and want to meet as many friends as possible. Solve the problem by considering various different schedules and picking the best one to optimize your goals.

Travel distances (in minutes):
Fisherman's Wharf to Embarcadero: 8.
Fisherman's Wharf to Golden Gate Park: 25.
Fisherman's Wharf to Mission District: 22.
Fisherman's Wharf to Union Square: 13.
Embarcadero to Fisherman's Wharf: 6.
.
.
.

CONSTRAINTS: You arrive at Fisherman's Wharf at 9:00AM. Barbara will be at Embarcadero from 12:15PM to 6:45PM. You'd like to meet Barbara for a minimum of 105 minutes. Ashley will be at Golden Gate Park from 9:30AM to 6:00PM. You'd like to meet Ashley for a minimum of 75 minutes. Ronald will be at 
Mission District from 2:30PM to 8:15PM. You'd like to meet Ronald for a minimum of 45 minutes. Robert will be at Union Square from 11:45AM to 9:00PM. You'd like to meet Robert for a minimum of 60 minutes.

Incorrect plan:
SOLUTION: You start at Fisherman's Wharf at 9:00AM. You travel to Golden Gate Park in 25 minutes and arrive at 9:25AM. You meet Ashley for 75 minutes from 9:25AM to 10:40AM. You travel to Union Square in 22 minutes and arrive at 11:02AM. You wait until 11:45AM. You meet Robert for 60 minutes from 
11:45AM to 12:45PM. You travel to Embarcadero in 10 minutes and arrive at 12:55PM. You meet Barbara for 105 minutes from 12:55PM to 2:40PM. You travel to Mission District in 20 minutes and arrive at 3:00PM. You meet Ronald for 45 minutes from 3:00PM to 3:45PM.

Errors with the above plan:
Had error: Invalid meeting time or location with step: 'You meet Ashley for 75 minutes from 9:25AM to 10:40AM'

Fixed plan: (Your response should start with 'SOLUTION:', and follow the same solution format as shown above.)

\end{lstlisting}

\subsubsection{Calendar Planning}
\begin{lstlisting}
You are an expert at scheduling meetings. You are given a few constraints on the existing schedule of each participant, the meeting duration, and possibly some preferences on the meeting time. Note there exists a solution that works with existing schedule of every participant. Produce only one meeting time, strictly adhering to the format shown in the examples below, without providing any extra information. Here are a few example tasks and solutions:

TASK: You need to schedule a meeting for Roger, Karen and Dorothy for half an hour between the work hours of 9:00 to 17:00 on Monday. 

Here are the existing schedules for everyone during the day: 
Roger's calendar is wide open the entire day.
Karen has meetings on Monday during 10:00 to 10:30, 11:30 to 12:00, 12:30 to 13:00, 14:00 to 15:00, 15:30 to 16:00; 
Dorothy is busy on Monday during 9:00 to 10:00, 10:30 to 11:00, 11:30 to 12:00, 12:30 to 13:00, 14:00 to 15:30, 16:00 to 17:00; 

You would like to schedule the meeting at their earlist availability.
Find a time that works for everyone's schedule and constraints. 
SOLUTION: Here is the proposed time: Monday, 11:00 - 11:30 

TASK: You ...

Query:
TASK: You need to schedule a meeting for Michelle, Steven and Jerry for one hour between the work hours of 9:00 to 17:00 on Monday. 

Here are the existing schedules for everyone during the day: 
Michelle has meetings on Monday during 11:00 to 12:00; 
Steven has blocked their calendar on Monday during 9:00 to 9:30, 11:30 to 12:00, 13:30 to 14:00, 15:30 to 16:00; 
Jerry has blocked their calendar on Monday during 9:00 to 9:30, 10:00 to 11:00, 11:30 to 12:30, 13:00 to 14:30, 15:30 to 16:00, 16:30 to 17:00; 

Find a time that works for everyone's schedule and constraints. 
SOLUTION: 

==============RESPONSE OF 4o-mini=================
Here is the proposed time: Monday, 12:00 - 13:00

==============Backprompt=================

You are an expert at scheduling meetings. You are given a few constraints on the existing schedule of each participant, the meeting duration, and possibly some preferences on the meeting time. Propose a different time to meet than the one provided below such that it meets as many specified constraints as possible. Note there exists a solution that works with existing schedule of every participant. Produce only one meeting time, strictly adhering to the format shown in the examples below, without providing any extra information. Here are a few example tasks and solutions:

TASK: You need to schedule a meeting for Roger, Karen and Dorothy for half an hour between the work hours of 9:00 to 17:00 on Monday. 

Here are the existing schedules for everyone during the day: 
Roger's calendar is wide open the entire day.
Karen has meetings on Monday during 10:00 to 10:30, 11:30 to 12:00, 12:30 to 13:00, 14:00 to 15:00, 15:30 to 16:00; 
Dorothy is busy on Monday during 9:00 to 10:00, 10:30 to 11:00, 11:30 to 12:00, 12:30 to 13:00, 14:00 to 15:30, 16:00 to 17:00; 

You would like to schedule the meeting at their earlist availability.
Find a time that works for everyone's schedule and constraints. 
SOLUTION: Here is the proposed time: Monday, 11:00 - 11:30 

TASK: You ...

Query:
TASK: You need to schedule a meeting for Michelle, Steven and Jerry for one hour between the work hours of 9:00 to 17:00 on Monday. 

Here are the existing schedules for everyone during the day: 
Michelle has meetings on Monday during 11:00 to 12:00; 
Steven has blocked their calendar on Monday during 9:00 to 9:30, 11:30 to 12:00, 13:30 to 14:00, 15:30 to 16:00; 
Jerry has blocked their calendar on Monday during 9:00 to 9:30, 10:00 to 11:00, 11:30 to 12:30, 13:00 to 14:30, 15:30 to 16:00, 16:30 to 17:00; 

Find a time that works for everyone's schedule and constraints. 
SOLUTION: 

Incorrect meeting time:
Here is the proposed time: Monday, 12:00 - 13:00

Errors with the above meeting time:
1. Jerry is busy on Monday between 11:30 and 12:30

Fixed meeting time: 
\end{lstlisting}

\subsection{Critic Extraction}\label{critic_extract}
\begin{lstlisting}
Assume you have a JSON with the following as a list. The list is the itinerary generated for each day of the trip. 

```
- llm\_response: An array containing objects with the following fields for each day of the trip:
  - day: Integer representing the day number.
  - people\_number: Integer representing the number of people.
  - current\_city: String representing the current city/location.
  - transportation: String representing transportation details (e.g., flight number, departure/arrival time).
  - breakfast: String representing breakfast arrangements.
  - attraction: String representing the attraction(s) for the day.
  - lunch: String representing lunch arrangements.
  - dinner: String representing dinner arrangements.
  - accommodation: String representing accommodation details.
```

Additionally, you have the following functions that you can use : 

1. get\_cost\_of\_transport(source, destination, mode-of-travel)['cost']
2. A valid restaurants dataframe with keys : 'Name','Average Cost','Cuisines','Aggregate Rating','City'
3. A valid attractions dataframe with keys : 'Name','Latitude','Longitude','Address','Phone','Website',"City"
4. A valid accomodations dataframe with keys : 'NAME','price','room type', 'house_rules', 'minimum nights', 'maximum occupancy', 'review rate number', 'city'
5. A valid flights dataframe with kyes : 'Flight Number', 'Price', 'DepTime', 'ArrTime', 'ActualElapsedTime','FlightDate','OriginCityName','DestCityName','Distance'


You task is to write 

A method called 'get\_total\_cost' which parses the json file and calculates the total cost of the trip. Ensure you take into account the cost of transport, breakfast, lunch, dinner, accomodation. Ensure that you take into account the number of people for which the itinerary has been made. Use the data within the json and the available dataframes and tools for estimating the total cost. Do not assume any other details. Implement the complete function with all the details. Return a float value representing the total cost. 

\end{lstlisting}

\end{document}